\documentclass[preprint,12pt]{elsarticle}

\usepackage{graphicx}
\usepackage{amssymb}

\usepackage{lineno}

\journal{}

\begin{document}

\begin{frontmatter}


\title{What's in my closet?: Image classification using fuzzy logic}



\author{Amina E. Hussein}

\address{University of Michigan, Ann Arbor, MI, 48109, USA}

\begin{abstract}
A fuzzy system was created in MATLAB to identify an item of clothing as a dress, shirt, or pair of pants from a series of input images. The system was initialized using a high-contrast vector-image of each item of clothing as the state closest to a direct solution. Nine other user-input images (three of each item) were also used to determine the characteristic function of each item and recognize each pattern. Mamdani inference systems were used for edge location and identification of characteristic regions of interest for each item of clothing. Based on these non-dimensional trends, a second Mamdani fuzzy inference system was used to characterize each image as containing a shirt, a dress, or a pair of pants. An outline of the fuzzy inference system and image processing techniques used for creating an image pattern recognition system are discussed. 
\end{abstract}

\begin{keyword}
Fuzzy sets \sep Computer vision \sep Image recognition


\end{keyword}

\end{frontmatter}


Fuzzy logic techniques have a wide range of applications in transforming complicated systems to tractable, linguistic representations. Some applications include process management and control, decision making, ecology, economics and pattern recognition and classification \cite{Zadeh, Phillis, image, Ferreira, Isermann}. Fuzzy logic allows a \textit{degree of membership} to be ascribed to data within a set. Using this, a \textit{fuzzy set} can be constructed, in which which each element has a membership value between 0\% and 100\% \cite{Zadeh, Mendel}. In this way, values within a set can be sorted by their degree of belonging, enabling classification beyond a binary \textit{yes} or \textit{no} representation. Therefore, fuzzy sets are very useful for pattern recognition, and have been employed for character and writing recognition, as well as voice recognition \cite{Hanmandlu, Melin}.\\

A fuzzy system was created to identify an item of clothing as a dress, shirt, or pair of pants, from a series of input images. The system was initialized using a high-contrast vector-image of each item of clothing as the state closest to a direct solution. Nine other user-input images (three of each item) were also used to determine the characteristic function of each item, and to train the system to recognize each pattern. In the first level of fuzzy image processing, a Mamdani inference system was used to locate edges in an input image \cite{Mamdani}. There characteristic regions of interest were then selected, and the left-most edge point was used to obtain characteristic curve of each item of clothing. Based on these non-dimensional trends, a second Mamdani fuzzy inference system was used to characterize each image as containing a shirt, a dress, or a pair of pants. \\

The motivation for this project was to develop a closet organization and outfit generation application based on user inputs of clothing images. This project was inspired by the 1995 movie \textit{Clueless}, in which the main character, Cher, has a computer program that classifies the items of clothing in her closet, and suggests pairs between items to form an outfit. However, despite twenty years of technological advancement since this visionary film was released, such a program does not yet exist! According to an article in Forbes Magazine in 2011, women make up 70\% of the e-commerce customer base, but the creation of web and mobile applications targeted to this demographic remains largely untapped \cite{women}. A closet organization application called \textit{Stylebook} currently sells on the Apple Store for $\$3.99$, indicating a market for these programs. However this application lacks image recognition and outfit matching capabilities, serving more as a image database than an equivalent to Cher's program. Therefore, beyond lending itself well to fuzzy logic techniques, the creation of a closet organization program also has financial incentives. \\

In this paper, an outline of the fuzzy inference system and image processing techniques used for creating an image pattern recognition system are discussed. An outline of the model is given in Section \ref{model}, and results are presented in Section \ref{results}. Limitations of the existing program as well as future extensions are discussed in Section \ref{conclusions}.

\section{Model}
\label{model}

\begin{figure}[thpb]
  \centering
    \includegraphics[width=80mm]{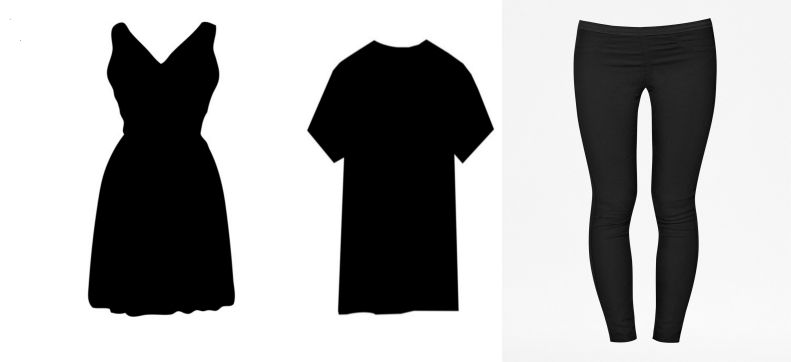}
  \caption{Template images of a dress, shirt and pair of pants used in the code.}
  \label{templates}
\end{figure}

To determine a characteristic function of three styles of clothing, shirt, dress or pants, three template images were upload. The template images were selected to be high-contrast and representative of the ideal form of each item. The images for the t-shirt and dress were vector images obtained as a free download from Tuts King, an online vector image resource. The template image for pants was obtained from the e-Commerce website French Connection, as it was more representative of women's pants than the vector images available, and thus would be more congruent with user-input data. These three template images are shown in Figure \ref{templates}. \\

\begin{figure}
  \centering
    \includegraphics[width=100mm]{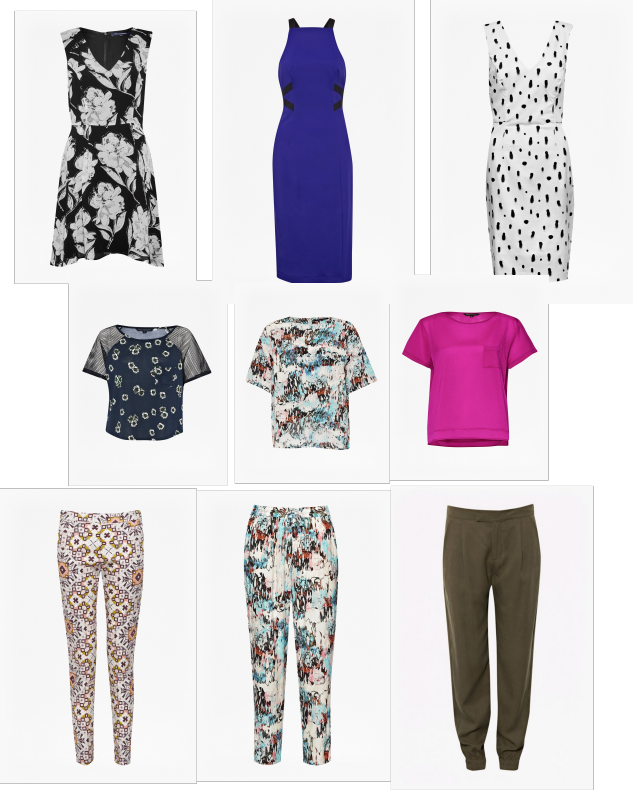}
  \caption{Real images of dresses, shirts and pants used in the code. Complex patterns were selected for each item to challenge the recognition capabilities of the program.}
  \label{real}
\end{figure}

For each clothing type, three images of real clothing obtained from the e-Commerce site French Connection were used to train the algorithm. For each item type, three patterns were chosen: floral print, geometric pattern and solid color. These items are shown in Figure \ref{real}. Patterned images were chosen so that the program could be built with diverse inputs, thus broadening it's real-world applications. All images and template images were scaled to the same dimensions of 1024 by 1536 pixels and loaded into the master function.\\

\begin{figure}
  \centering
    \includegraphics[width=100mm]{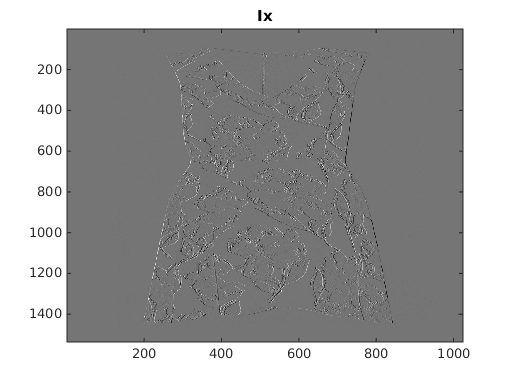}
  \caption{Gradient along the x-axis of the input image of a floral dress.}
  \label{gradients}
\end{figure}

The first step in image processing was to determine the edges of each image. This was done using an edge detection Fuzzy Inference System (FIS). The development of this FIS followed a Mathworks tutorial for fuzzy image processing in MATLAB. This program calculates the effective luminance of each pixel in a nxnx3 image. The image was then converted to grey-scale and to a double array. For each image, the intensity range was scaled to exist between 0 and 1, where 0 corresponded to black and 1 to white. A gradient between -1 and 1 was set up in the x-axis and y-axis, and convolved with the image to obtain gradients in each direction, as shown for the x-direction in Figure \ref{gradients}. These gradient profiles were the input values to the FIS.\\

\begin{figure}
  \centering
    \includegraphics[width=80mm]{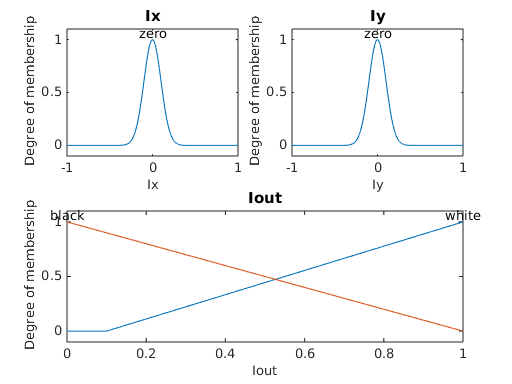}
  \caption{Membership functions of input variables $Ix$ and $Iy$ and output variable $output$ used in \textit{edgeFIS}.}
  \label{edgeFIS}
\end{figure}

The FIS for edge detection had two input values, the gradients along the x-axis and y-axis ($Ix$ and $Iy$), each with a Gaussian membership function centered at zero. The standard deviation of the membership function was tuned for the images, where a smaller value made the edge detection more sensitive and vice versa. For the template images, a standard deviation of 0.1 for both $Ix$ and $Iy$ was found to give the best results. A standard deviation of 0.3 was optimal for the user input images by allowing more subtle intensity changes to be detected. This was particularly useful for resolving white items on a white background, such as the polka dot dress shown in Figure \ref{real}. The output variable, $output$, had bounds between 0 and 1, and consisted of two triangular membership functions corresponding to $white$ and $black$. The input and output membership functions using on the template images are shown in Figure \ref{edgeFIS}. Two rules were defined: 

\begin{itemize}
\item[1] If Ix is zero and Iy is zero then Iout is white
\item[2] If Ix is not zero or Iy is not zero then Iout is black
\end{itemize}

\begin{figure}
  \centering
    \includegraphics[width=120mm]{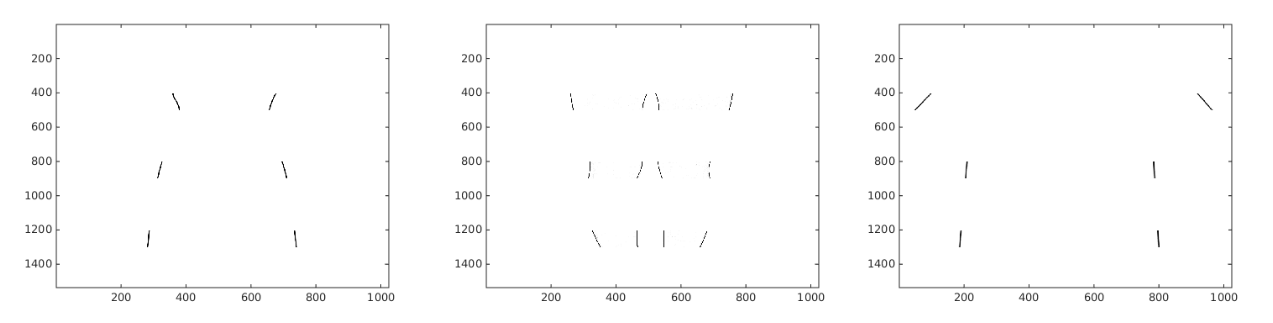}
  \caption{Regions of interest (ROIs) defined for each clothing item from the template image to capture defining features.}
  \label{ROI}
\end{figure}

The output from this FIS was allocated to a matrix with the same dimensions as the initial image and returned to the image processing subroutine. This data was then sent to a matrix function where the edge values were normalized to set the value of a white pixel to 1. Three region of interest (ROIs) were chosen for each item of clothing. These ROIs were set at rows 400, 800 and 1200, (corresponding to ROI one, two and three) with a pixel spread of 100. The use of ROIs enables greater detection precision by limiting the scanning area in each image to only capture defining features. As well, choosing three regions of 3x1024 pixels greatly increased the speed and efficiency of the code. These ROIs were selected for their distinct features for each item type, as shown in Figure \ref{ROI}.\\

Once ROIs were determined for each image, the leftmost non-white pixel (i.e. with intensity less than 0.98), corresponding to the beginning of the outline of the item of clothing from the left, was found. This was done so that a characteristic curve of each item could be found, independent of internal patterns of intensity fluctuations. A threshold of 0.98 was found to be optimal for resolving all important features from user-input images. The first and last fifty columns of each image were set to 1 (white) to remove any irrelevant artifacts from the outside of images (such a boarder). This program was applied to both template and user-input images along the left side only so that a single value function corresponding to the variation of the object could be obtained. \\

The most important output of this function was the column values corresponding to the first non-zero energy in each row. Using this output, a plot of column values versus row index provided an output function representative of each item of clothing. A first-order polynomial was fit to intensity values in ROI one. The mean intensity value in ROIs one and two was calculated and subtracted from each other. The choice of these variables is explained in Section \ref{results}. This image processing pipeline as completed for each of the twelve training images. \\

At the end of the analysis, twelve values of the slope of ROI one (dubbed \textit{m1}) and twelve values of the difference between means values in ROI one and two (dubbed \textit{meanVal}) were returned to the master function. These values were separated according to the known items: four values of $m1$ for dresses (\textit{m1\_ dresses}), pants (\textit{m1\_ pants}) and shirts (\textit{m1\_ shirt}), and four values of $meanVal$ for dresses (\textit{diff\_ dresses}) and pants (\textit{diff\_ pants}) . The $meanVal$ for shirts were neglected because they could resolved using $m1$. These values were used to construct the FIS \textit{identify} for object recognition, with $m1$ and $meanVal$ as the input variables. The ranges of $m1$ and $meanVal$ were dictated by their minimum and maximum values for training photos, and later expanded to [-100,100] and [-1000,1000], respectively, to maximize the detection range. . \\

\begin{figure}
  \centering
    \includegraphics[width=85mm]{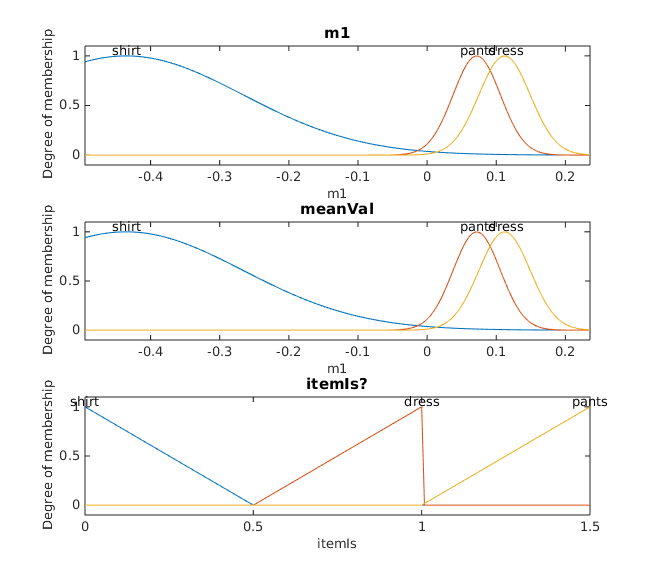}
  \caption{Membership functions for the input variables $m1$ and $meanVal$ and output variable $itemIS$ used in the object recognition FIS $identify$.}
  \label{FIS}
\end{figure}

Three Gaussian membership functions were created for the first variable $m1$: \textit{shirt}, with standard deviation determined from std(\textit{m1 \_ shirt}) and centered at mean(\textit{m1 \_ shirt}), \textit{pants}, with standard deviation std(\textit{m1 \_ pants}) and centered at mean(\textit{m1 \_ pants}) and $dress$, with std(\textit{m1 \_ dresses}) and centered at mean(\textit{m1 \_ dresses}). Two Gaussian membership functions were created for $meanVal$: $dress$ and $pants$, centered at mean(\textit{diff\_ dresses}) and mean(\textit{diff\_ pants}), respectively, each with a standard deviation equal to their own mean, to make a distinction between positive and negative mean value differences. A single output variable was defined, named $itemIs$, which ranged between 0 and 1.5. This output consisted of three triangular membership functions: $shirt$ ranging between 0 and 0.5, $dress$ between 0.5 and 1, and $pants$ between 1 and 1.5. The membership functions of the input and output variables in the FIS \textit{identify} are shown in Figure \ref{FIS}. Four rules were defined for this FIS:

\begin{itemize}
\item[1] If m1 is shirt then itemIs is shirt
\item[2] If m1 is not shirt and meanVal is dress then itemIs is dress
\item[3] If m1 is not shirt and meanVal is pants then itemIs is pants
\item[4] If m1 is shirt and meanVal is dress then itemIs is dress
\end{itemize}

Using this FIS, shirts were evaluated from the output as items with values less than 0.5, dresses had values between 0.5 and 1.0, and pants had values between 1.0 and 1.5. Therefore, for a given image, the matlab script would output the figure value from the input array corresponding to its index in the input array, and state if the image contained a dress, shirt or pair of pants. Using this trained system, any new image could be introduced to the system, processed to obtain its slope in the first region of interest and difference between the average intensity values in its first and second regions of interest, and the script would identify the object. 

\section{Results}
\label{results}

\begin{figure}
  \centering
    \includegraphics[width=120mm]{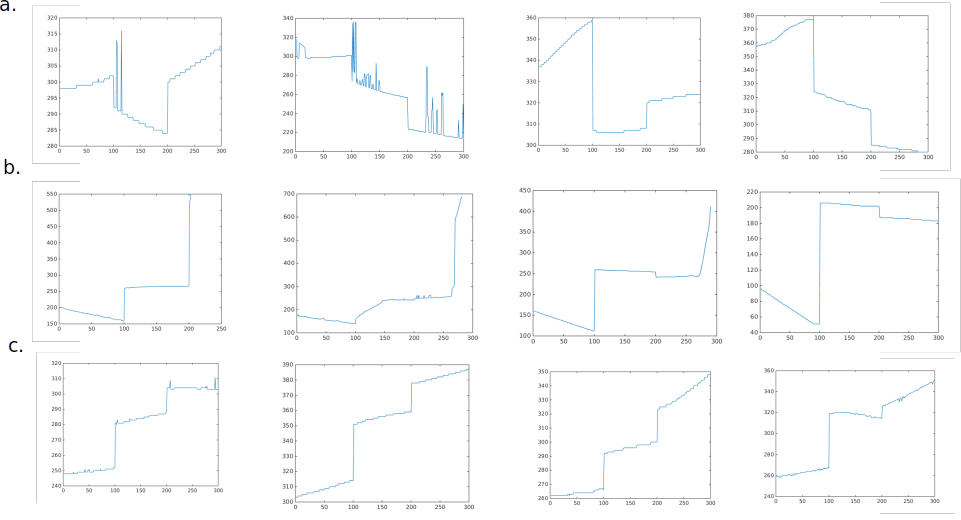}
  \caption{Characteristic functions obtained from twelve training images of dresses (a), shirts (b) and pants (c) from which each item could be determined.}
  \label{char}
\end{figure}

The greatest challenge in this project was determining a non-dimensional representation of each item of clothing (shirt, dress or pants) from the left-most edge intensity output of each training image obtained from the edge detection. The output function was observed as a single function (containing ROIs one, two and three together), and in parts with each ROI distinct. Fourier transforms and curve fitting did not give useful outputs that could be generalized for each item. However, from observing the slopes in each ROI for each item, a pattern became obvious: shirts were characterized by a negative slope in ROI one, where as pants and dresses both had positive slopes. This enabled easy separation of shirt inputs, but was useless to resolve dresses versus pants. The slopes in ROIs two and three did not provide any more information; ROI two did not differ significantly in any of the images, and ROI three provided information similar to that obtained in ROI one. However, from observing the function trends for each training image, as shown in Figure \ref{char}, it was found that pants and dresses could be differentiated by the difference between their mean intensity values in ROIs one and two. As well, it was found that shirts and pants both had a negative mean value difference between ROI one and two. Therefore, if an item was defined as a shirt, but had a positive mean value difference, it could be correctly identified as a dress. These signatures were used to construct the four rules in the identification FIS outlined in Section \ref{model}\\

\begin{figure}
  \centering
    \includegraphics[width=120mm]{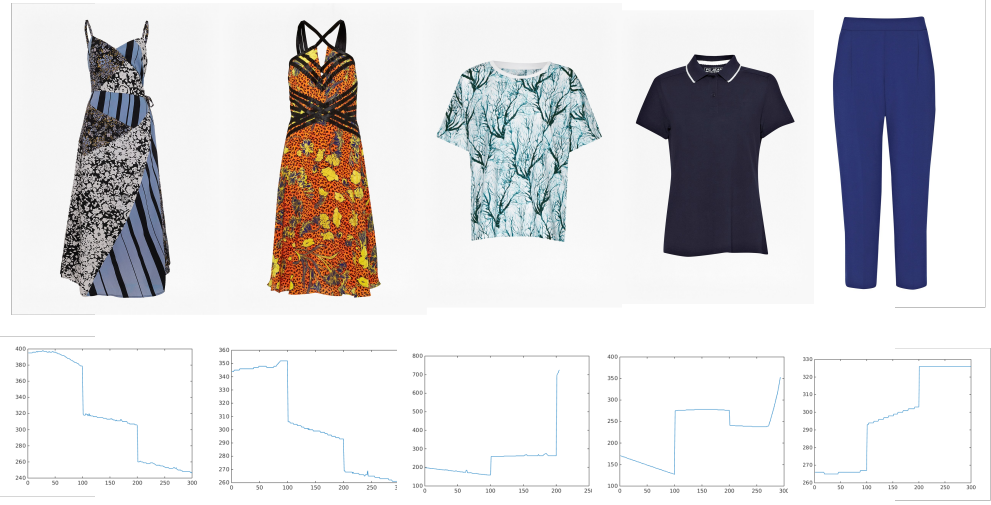}
  \caption{Five test items of clothing and their characteristic functions.}
  \label{tests}
\end{figure}

The FIS was trained using using the 12 images shown in Figure \ref{real}. It was then tested using five test cases, shown in Figure \ref{tests} with their corresponding characteristic functions. The program correctly identified the item of clothing in all five of the test cases. Therefore, the program is deemed successful, at least for a small pool of test cases. \\

The total run-time of the program totaled 184.185 seconds for the uploading, processing and testing of 17 images with dimensions 1064$\times$1536. The majority of the processing time was spent in image processing, specifically in locating the edges of the image using the edge detection FIS. Further work could be done to improve the speed and efficiency of this program to enable the collection of real-time object identification results.

\section{Conclusions}
\label{conclusions}

In this project, an object identification FIS was created, based on an edge detection FIS and image processing to determine a single value characteristic function of three clothing types: a shirt, a dress and a pair of pants. This program was able to successfully determine is an item was a shirt in all five test cases, and therefore is deemed successful. \\

Some limitations of this program are that the user-input images must have high contrast backgrounds. As well, the sensitivity of the edge FIS may need to be fine tuned for images taken with a mobile phone, as is the intended application of this program. For more reliability and detection accuracy, the program should be trained with greater than 12 images. One way to do this would be by generating a public forum in which users could upload their own items to train the program. \\

This program could be extended to detect clothing patterns, such as stripes, polka-dots and floral prints, by finding a non-dimensional characteristic function for each of these common patterns. A pattern FIS could then be constructed. As well, once the images are labeled as their clothing type, a fuzzy system could be created to suggest outfits, based on item rules (such as, pants cannot be worn with pants, shirts are worn with pants, dresses are worn alone). This could be extended to take in factors such as the weather, occasion and the users mood. With these advancements, this program would be even more desirable than Cher's from \textit{Clueless}!

\section*{Acknowledgement}

This project was completed while the author was at Purdue University. The author would like to acknowledge Professor Lefteri H. Tsoukalas for introduction to methods in fuzzy logic, and Dr. Wylie Stroberg for fruitful discussions and support of this project.

\bibliography{NUCL570_report}

\begin{thebibliography}{10}

\bibitem{Zadeh}
G.~J.~Klir L.~A.~Zadeh and B.~Yuan.
\newblock {\em Fuzzy Sets, Fuzzy Logic, and Fuzzy Systems: Selected Papers by
  Lofti A. Zadeh}.
\newblock World Scientific Publishing Co., Inc., River Edge, NJ, 1996.

\bibitem{Phillis}
L.~A.~Andriantiatsaholiniaina Y.~A.~Phillis.
\newblock Sustainability: an ill-defined concept and its assessment using fuzzy
  logic.
\newblock {\em Ecological Economics}, 2001.

\bibitem{image}
I.~Nedelijovic.
\newblock Image classification based on fuzzy logic.
\newblock {\em International Archives of the Photogrammetry, Remote Sensing and
  Spatial Information Sciences}, 2004.

\bibitem{Ferreira}
G.~Spiazzi A.~A.~Ferreira, J. A.~Pomilio et~al.
\newblock Energy management fuzzy logic supervisory for electric vehicle power
  supplies system.
\newblock {\em IEEE Transactions on Power Electronics}, 2008.

\bibitem{Isermann}
R.~Isermann.
\newblock On fuzzy logic applications for automatic control, supervision, and
  fault diagnosis.
\newblock {\em IEEE Transactions on Systems, Man, and Cybernetics - Part A:
  Systems and Humans}, 1998.

\bibitem{Mendel}
J.~M. Mendel.
\newblock Fuzzy logic systems for engineering: A tutorial.
\newblock {\em Proceedings of the IEEE}, 1995.

\bibitem{Hanmandlu}
S.~Chakraborty M.~Hanmandlu, K. R. Murali~Mohan et~al.
\newblock Unconsstrained handwritten character recognition based on fuzzy
  logic.
\newblock {\em Pattern Recognition}, 2003.

\bibitem{Melin}
P.~Melin and O.~Castillo.
\newblock {\em Voice Recognition with Neural Networks, Fuzzy Logic and Genetic
  Algorithms. In: Hybrid Intelligent Systems for Pattern Recognition Using Soft
  Computing. Studies in Fuzziness and Soft Computing}.
\newblock Springer, Berlin, Heidelberg, 2005.

\bibitem{Mamdani}
E~H. Mamdani and S.~Assilian.
\newblock An experiment in linguistic synthesis with a fuzzy logic controller.
\newblock {\em International Journal of Man-Machine Studies}, 1975.

\bibitem{women}
Steve Denning.
\newblock Women are the rock fuel of ecommerce.
\newblock {\em Forbes Magazine}, 2011.

\end{thebibliography}


\begin{thebibliography}{1}

\bibitem{women}
Steve Denning.
\newblock Women are the rock fuel of ecommerce.
\newblock {\em Forbes Magazine}, 2011.

\bibitem{image}
I.~Nedelijovic.
\newblock Image classification based on fuzzy logic.
\newblock {\em International Archives of the Photogrammetry, Remote Sensing and
  Spatial Information Sciences}, 2004.

\end{thebibliography}
\bibliographystyle{unsrt}

\addtolength{\textheight}{-12cm}   





\end{document}